\documentclass{article}

\usepackage[preprint,nonatbib]{neurips_2021}

\usepackage[utf8]{inputenc} % allow utf-8 input
\usepackage[T1]{fontenc}    % use 8-bit T1 fonts
\usepackage{hyperref}       % hyperlinks
\usepackage{url}            % simple URL typesetting
\usepackage{booktabs}       % professional-quality tables
\usepackage{amsfonts}       % blackboard math symbols
\usepackage{nicefrac}       % compact symbols for 1/2, etc.
\usepackage{microtype}      % microtypography

\usepackage{graphicx}
\usepackage{amsmath}
\usepackage{amssymb}
\usepackage[shortlabels]{enumitem}
\usepackage[font={small}]{caption}
\usepackage[multiple]{footmisc}
\usepackage{multirow}
\usepackage{booktabs}
\usepackage[T1]{fontenc}
\usepackage[dvipsnames]{xcolor}
\usepackage{subfig}
\usepackage{authblk}        % multiple authors

\newcommand\jb[1]{\textcolor{blue}{[JB: #1]}}

\newcommand\bert{\textsc{BERT}}

\newcommand\tfive{\textsc{T5}}
\newcommand\unifiedqa{\textsc{UnifiedQA}}
\newcommand\comment[1]{}

\newcommand\softmax{$\mathrm{softmax}$}
\newcommand\relu{$\mathrm{ReLU}$}

\usepackage{xcolor}
\definecolor{maroon}{cmyk}{0, 0.87, 0.68, 0.32}
\definecolor{halfgray}{gray}{0.55}
\definecolor{ipython_frame}{RGB}{207, 207, 207}
\definecolor{ipython_bg}{RGB}{247, 247, 247}
\definecolor{ipython_red}{RGB}{186, 33, 33}
\definecolor{ipython_green}{RGB}{0, 128, 0}
\definecolor{ipython_cyan}{RGB}{64, 128, 128}
\definecolor{ipython_purple}{RGB}{170, 34, 255}

\usepackage{listings}
\lstdefinelanguage{iPython}{
    morekeywords={access,and,break,class,continue,def,del,elif,else,except,exec,finally,for,from,global,if,import,in,is,lambda,not,or,pass,print,raise,return,try,while},%
    morekeywords=[2]{abs,all,any,basestring,bin,bool,bytearray,callable,chr,classmethod,cmp,compile,complex,delattr,dict,dir,divmod,enumerate,eval,execfile,file,filter,float,format,frozenset,getattr,globals,hasattr,hash,help,hex,id,input,int,isinstance,issubclass,iter,len,list,locals,long,map,max,memoryview,min,next,object,oct,open,ord,pow,property,range,raw_input,reduce,reload,repr,reversed,round,set,setattr,slice,sorted,staticmethod,str,sum,super,tuple,type,unichr,unicode,vars,xrange,zip,apply,buffer,coerce,intern},%
    sensitive=true,%
    morecomment=[l]\#,%
    morestring=[b]',%
    morestring=[b]",%
    morestring=[s]{'''}{'''},% used for documentation text (mulitiline strings)
    morestring=[s]{"""}{"""},% added by Philipp Matthias Hahn
    morestring=[s]{r'}{'},% `raw' strings
    morestring=[s]{r"}{"},%
    morestring=[s]{r'''}{'''},%
    morestring=[s]{r"""}{"""},%
    morestring=[s]{u'}{'},% unicode strings
    morestring=[s]{u"}{"},%
    morestring=[s]{u'''}{'''},%
    morestring=[s]{u"""}{"""},%
    literate=
    {á}{{\'a}}1 {é}{{\'e}}1 {í}{{\'i}}1 {ó}{{\'o}}1 {ú}{{\'u}}1
    {Á}{{\'A}}1 {É}{{\'E}}1 {Í}{{\'I}}1 {Ó}{{\'O}}1 {Ú}{{\'U}}1
    {à}{{\`a}}1 {è}{{\`e}}1 {ì}{{\`i}}1 {ò}{{\`o}}1 {ù}{{\`u}}1
    {À}{{\`A}}1 {È}{{\'E}}1 {Ì}{{\`I}}1 {Ò}{{\`O}}1 {Ù}{{\`U}}1
    {ä}{{\"a}}1 {ë}{{\"e}}1 {ï}{{\"i}}1 {ö}{{\"o}}1 {ü}{{\"u}}1
    {Ä}{{\"A}}1 {Ë}{{\"E}}1 {Ï}{{\"I}}1 {Ö}{{\"O}}1 {Ü}{{\"U}}1
    {â}{{\^a}}1 {ê}{{\^e}}1 {î}{{\^i}}1 {ô}{{\^o}}1 {û}{{\^u}}1
    {Â}{{\^A}}1 {Ê}{{\^E}}1 {Î}{{\^I}}1 {Ô}{{\^O}}1 {Û}{{\^U}}1
    {œ}{{\oe}}1 {Œ}{{\OE}}1 {æ}{{\ae}}1 {Æ}{{\AE}}1 {ß}{{\ss}}1
    {ç}{{\c c}}1 {Ç}{{\c C}}1 {ø}{{\o}}1 {å}{{\r a}}1 {Å}{{\r A}}1
    {€}{{\EUR}}1 {£}{{\pounds}}1
    {^}{{{\color{ipython_purple}\^{}}}}1
    {=}{{{\color{ipython_purple}=}}}1
    {+}{{{\color{ipython_purple}+}}}1
    {*}{{{\color{ipython_purple}$^\ast$}}}1
    {/}{{{\color{ipython_purple}/}}}1
    {+=}{{{+=}}}1
    {-=}{{{-=}}}1
    {*=}{{{$^\ast$=}}}1
    {/=}{{{/=}}}1,
    literate=
    *{-}{{{\color{ipython_purple}-}}}1
     {?}{{{\color{ipython_purple}?}}}1,
    identifierstyle=\color{black}\ttfamily,
    commentstyle=\color{ipython_cyan}\it\ttfamily,
    stringstyle=\color{ipython_red}\ttfamily,
    keepspaces=true,
    emph={forward,backward},                    % name of the functions
    emphstyle=\color{blue}\ttfamily,
    showspaces=false,
    showstringspaces=false,
    frame=tb,
    numbers=left,
    numberstyle=\tiny,
    basicstyle=\footnotesize,
    keywordstyle=\color{ipython_green}\ttfamily,
}
\usepackage{float}     % more table positioning options
\restylefloat{table}
\usepackage[leftcaption]{sidecap}  % for placing captions on side

\title{Memory-efficient Transformers via Top-$k$ Attention}

\author{
\textbf{Ankit Gupta}$^1$\thanks{majority of work done while author was part of IBM AI Residency program. \\ \phantom{x}\hspace{3ex}{\scriptsize {\tt Email:\ \{ankitgupta.iitkanpur,\;guyd1995,\;ishayahu156,\;davidciprut\}@gmail.com,\;joberant@cs.tau.ac.il}}} \ \quad
\textbf{Guy Dar}$^2$ \ \quad
\textbf{Shaya Goodman}$^2$ \ \quad
\textbf{David Ciprut}$^2$ \ \quad
\textbf{Jonathan Berant}$^{2,3}$\\
  $^1$IBM Research \qquad
  $^2$Tel Aviv University \qquad
  $^3$Allen Institute for AI
}

\begin{document}

\maketitle

\begin{abstract}

Following the success of dot-product attention in Transformers, numerous approximations have been recently proposed to address its quadratic complexity with respect to the input length.
While these variants are memory and compute efficient, it is not possible to directly use them with popular pre-trained language models trained using vanilla attention, without an expensive corrective pre-training stage. 
In this work, we propose a simple yet highly accurate approximation for vanilla attention. We process the queries in chunks, and for each query, compute the top-$k$ scores with respect to the keys. Our approach offers several advantages: (a) its memory usage is linear in the input size, similar to linear attention variants, such as Performer and RFA (b) it is a drop-in replacement for vanilla attention that does not require any corrective pre-training, and (c) it can also lead to significant memory savings in the feed-forward layers after casting them into the familiar query-key-value framework. 
We evaluate the quality of top-$k$ approximation for multi-head attention layers on the Long Range Arena Benchmark, and for feed-forward layers of T5 and UnifiedQA on multiple QA datasets. We show our approach leads to accuracy that is nearly-identical to vanilla attention in multiple setups including training from scratch, fine-tuning, and zero-shot inference.

\end{abstract}

\section{Introduction}\label{sec:intro}

The Transformer architecture \cite{vaswani2017attention} has been successful in a wide range of natural language processing tasks, including machine translation \cite{edunov2018understanding}, language modeling \cite{roy2020efficient}, question-answering \cite{karpukhin2020dense}, and many more. Transformers pre-trained on large amounts of text with a language modeling (LM) objective, have become the standard in NLP, exhibiting surprising amounts of linguistic and world knowledge \cite{peters2018elmo, devlin2018bert, petroni2019language, hewitt2019structural,Roberts2020t5kb}.

The contextualizing component of the Transformer is the attention layer where all positions in an input sequence of length $L$ aggregate information from the entire sequence in parallel. At its core, given $L$ query, key, and value vectors $Q, K, V$ respectively, the \textit{dot-product attention} function outputs
%\footnote{Usually, the term is $\mathrm{softmax}(QK^\top / \sqrt{d})V$ but $\sqrt{d}$ can be dropped via scaling of queries and keys.} 
$\mathrm{softmax}(QK^\top)V$ where the \softmax{} function is applied row-wise on the matrix $QK^\top \in \mathbb{R}^{L \times L}$ of similarity scores of the query-key pairs, leading to an expensive $\Omega(L^2)$ memory requirement.

To alleviate this, past work proposed approximation methods for the computation of $\mathrm{softmax}(QK^\top)$. One major line of research focused on \textit{sparse attention} variants, where only a few similarity scores are computed per query, and the rest are ignored. Methods differ by which query-key pairs are selected \cite{child2019generating, ye2019bp, qiu2019blockwise, roy2020efficient, kitaev2020reformer, beltagy2020longformer,gupta2020gmat,vyas_et_al_2020}. 
A second line of research explored \textit{dense} variants  \cite{katharopoulos2020transformers,Wang2020LinformerSW,bello2021lambdanetworks,tay2020sparse} (cf.\ \cite{tay2020efficient} for a survey). For example, instead of computing the attention scores exactly for only a small number of query-key pairs, \cite{Choromanski2020RethinkingAW} compute an approximation of scores for all pairs.

\begin{figure}[t]
    \centering
    \includegraphics[scale=0.54]{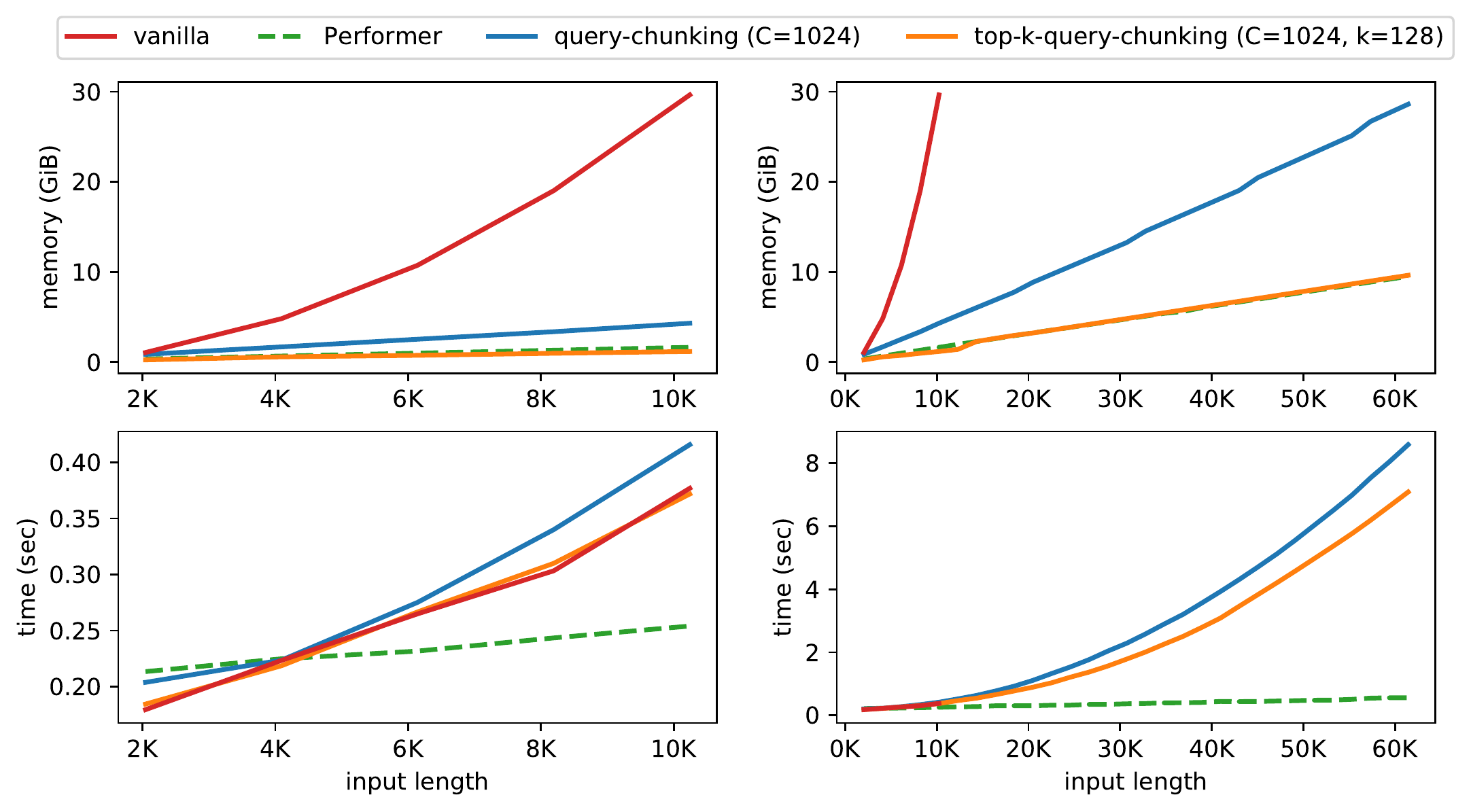}
    \setlength{\belowcaptionskip}{-15pt}   
    \caption{Memory and time required for a forward and backward pass on a single \bert{}-base multi-head self-attention layer with causal masking on short (left) and long (right) inputs. Details are in \S\ref{sec:benchmark}.} % top-$k$ attention reports the same memory-usage as linear attention variants such as the ``Performer''.}
\label{figure:mha}
\end{figure}

In this work, we adopt the sparse attention approach, but rather than \emph{approximating} the $k$ most similar key vectors per query vector, we compute this quantity \emph{exactly}. Specifically, we propose \textit{top-$k$ attention} where, for each query vector, we only keep its $k$ largest similarity scores with respect to the $L$ keys, where $k \ll L$. We show that top-$k$ attention can be implemented in a memory-efficient manner by (a)
chunking the query vectors when computing the output one chunk at a time, when computing $\mathrm{softmax}(QK^\top)V$, and (b) a custom implementation of the forward and backward pass that does not require caching activations while processing chunks in the forward pass.

Compared to prior methods, top-$k$ attention has multiple attractive properties:
\begin{itemize}[leftmargin=*,topsep=0pt,itemsep=4pt,parsep=0pt]
\item Top-$k$ attention has the same memory footprint as Performer \cite{Choromanski2020RethinkingAW}, a state-of-the-art attention variant with linear time and memory complexity, on very long inputs (orange curve, Fig.~\ref{figure:mha}, top-right), while being as fast as vanilla attention, and even faster than linear variants on inputs of length up to $4K$ (Figure~\ref{figure:mha}, bottom-left). This allows us, e.g., to train a typical 12-layer Transformer decoder over $32K$-long inputs on a $30$GiB GPU (Figure~\ref{figure:base}). 

\item Top-$k$ attention also reduces memory consumption in Transformer feed-forward layers, by casting this layer into the familiar query-key-value framework using \relu{} instead of the row-wise \softmax{} \cite{sukhbaatar2020augmenting}.
This is specifically appealing in models such as T5 \cite{raffel2019exploring} and GPT-3 \cite{Brown2020LanguageMA}, where for short inputs, the memory consumption is dominated by the feed-forward layers, as
the number of keys, corresponding to the feed-forward hidden dimension size, is as large as $65K$. Conversely, methods that rely on random feature approximations of attention, such as Performer \cite{Choromanski2020RethinkingAW} and RFA \cite{peng2021random} do not admit an efficient approximation for the \relu{} activation \cite{yehudai2019power}.

\item Top-$k$ attention is a highly accurate approximation to vanilla attention and is a \textit{plug-and-play} replacement at both multi-head attention and feed-forward layers of a Transformer. This is unlike past attention variants \cite{katharopoulos2020transformers,Choromanski2020RethinkingAW,peng2021random} that require an expensive corrective pre-training stage to adjust model weights to the new variant, which can be prohibitive for large models. We show top-$k$ attention can replace vanilla attention in a zero-shot inference setup and at fine-tuning time without any corrective pre-training.
\end{itemize}

We extensively evaluate top-$k$ attention on a wide range of tasks and demonstrate its mentioned advantages. Training from scratch, we show top-$k$ attention performs as well as vanilla self-attention on Long Range Arena, a benchmark dedicated to evaluating the ability of transformers to handle long sequences, and in a language modeling task (WikiText-103). Second, we show top-$k$ attention can be used as a drop-in replacement for vanilla attention 
at inference time without any additional training at the feed-forward layer of the UnifiedQA model \cite{2020unifiedqa} on 12 different question answering (QA) datasets, reducing the number of keys used per query by more than 99\%. Last, we show top-$k$ attention obtains similar performance to vanilla attention on a wide range of QA tasks when fine-tuning T5 \cite{raffel2019exploring}, without the need for any corrective pre-training.

Overall, our results demonstrate that top-$k$ attention is a simple and effective method for dramatically reducing the memory footprint of Transformers without loss of accuracy that can allow resource-constrained researchers enjoy the benefits of large pre-trained Transformer-based models. Our code is available at \url{https://github.com/ag1988/top_k_attention}.

\section{Efficient Transformer through Top-$k$ Attention}\label{sec:model}

In this section, we briefly review the Transformer architecture, its sparse approximations, and show how to cast the feed-forward layer into the query-key-value framework (\S\ref{subsec:att_in_transformers}). We then describe top-$k$ attention and our memory-efficient implementation for it (\S\ref{subsec:top-k-att}).

\subsection{Attention in Transformers}
\label{subsec:att_in_transformers}

%Before formally describing our experimental set-up we recap some relevant definitions from \cite{vaswani2017attention}. 
A Transformer \cite{vaswani2017attention} is a stack of layers each consisting of multi-head attention and feed-forward sub-layers. Its contextualizing component is the multi-head attention defined as follows.

\paragraph{Multi-head Attention} Given a query $Q \in \mathbb{R}^{L_Q \times d}$, key $K \in \mathbb{R}^{L_K\times d}$ and value $V \in \mathbb{R}^{L_K\times d}$, the output $\in \mathbb{R}^{L_Q \times d}$ of dot-product attention is defined as:
\begin{equation}\label{eqn:attention}
\mathrm{Attention}(Q,K,V) = \mathrm{row}{\text -}\mathrm{softmax}\left(\frac{QK^\top}{\lambda}\right)V,
\end{equation}
where $\lambda$ is an optional temperature typically fixed as $\sqrt{d}$.\footnote{we omit this in rest of our presentation as $Q$ can be scaled by $1 / \lambda$ beforehand.} In multi-head attention, for a given number of heads $h$, instead of computing a single attention output with $d_{\text{model}}$ dimensional queries, keys and values, these are linearly projected down in parallel $h$ times to $d = d_{\text{model}} / h$ dimensions, using different learned projection matrices. $\mathrm{Attention}$ is applied to each of the $h$ new queries, keys and values, yielding $d$ dimensional outputs which are concatenated and again projected to obtain a $d_\text{model}$-dimensional output.

\paragraph{Sparse approximations} The attention function (Eq. \ref{eqn:attention}) requires the computation of $QK^\top$ containing $L_Q\cdot L_K$ entries and can be expensive for long sequences ($L_Q$ and $L_K$ are typically the sequence length). To alleviate this issue, \textit{sparse attention} variants \cite{child2019generating, qiu2019blockwise, kitaev2020reformer, beltagy2020longformer, gupta2020gmat} relax this requirement and compute only a few entries of $QK^\top$, masking out the rest. For a binary mask $B \in \{0,-\infty\}^{L_Q\times L_K}$,
\begin{equation}\label{eqn:sparse-attn}
\mathrm{SparseAttention}(Q,K,V,B) = \mathrm{row}{\text -}\mathrm{softmax}\left(QK^\top + B\right)V .
\end{equation}

The sparsity of $B$ can be leveraged via customized implementations of matrix product \cite{child2019generating,beltagy2020longformer} and, thus Eq. \ref{eqn:sparse-attn} can be significantly cheaper to compute compared to Eq. \ref{eqn:attention}.

\paragraph{Feed-forward as attention} In the feed-forward layer, a $1$-hidden layer fully-connected network is applied identically to every input token.
%\footnote{For brevity, we use the word \textit{token} to refer both to the input token and its contextualized representation.} 
As observed in past work \cite{sukhbaatar2020augmenting,Shazeer2020GLUVI,geva2020transformer}, a feed-forward layer can be cast into the query-key-value framework as:
\begin{equation}\label{eqn:ff}
\mathrm{FF}_{K,V}(Q) = \mathrm{ReLU}\left(QK^\top\right){V}.
\end{equation}
In this case, the queries $Q \in \mathbb{R}^{L_Q\times d_{\text{model}}}$ are the inputs to the layer with $L_Q$ tokens, similar to self-attention. However, the keys $K = W_K \in \mathbb{R}^{L_K\times d_{\text{model}}}$ and values $V = W_V \in \mathbb{R}^{L_K\times d_{\text{model}}}$ are learned parameters that are independent of the input.
The number of keys $L_K$ here is known as the \emph{feed-forward dimension} and can be as large as $65K$ for wide models such as T5 \cite{raffel2019exploring} and GPT-3 \cite{Brown2020LanguageMA}. In the common case where the input sequences are relatively short, memory consumption is dominated by the feed-forward sub-layer and not the self-attention sub-layer.

Unlike top-$k$ attention, past approaches for approximating attention are incompatible with feed-forward layers. Most approximate attention variants, such as Sparse Transformer \cite{child2019generating}, LongFormer \cite{beltagy2020longformer}, BigBird \cite{zaheer2020bigbird}, Sinkhorn attention \cite{tay2020sparse}, rely on a \emph{locality bias} in sequences, where the key vectors indexed close to each other in $K$ are assumed to have similar representations. This is irrelevant for keys in a feed-forward layer, which are permutation-equivariant and do not have any local structure. Dense attention variants relying on random fourier features for approximating the $\mathrm{softmax}$ function are also not applicable, since it is known that \relu{} cannot be approximated using such features \cite{yehudai2019power}.

\subsection{Top-$k$ Attention} 
\label{subsec:top-k-att}

In this work we propose top-$k$ attention, where for each query, we mask out all but its $k$ largest dot products with the keys, that is, in each row of $QK^\top$ we only keep its $k$ largest elements and mask out the rest:
\begin{equation}\label{eqn:topk}
\mathrm{top}{\text -}k{\text -}\mathrm{Attention}(Q,K,V) = \mathrm{activation}\left(\mathrm{top}{\text -}k(QK^\top)\right)V,
\end{equation}
where $\mathrm{activation}$ can be $\mathrm{softmax}$, \relu{}, or any other activation, and $\mathrm{top}{\text -}k(QK^\top)$ denotes a sparse matrix consisting only of the row-wise top-$k$ elements of $QK^\top$. A na\"ive approach for computing $\mathrm{top}{\text -}k(QK^\top)$ would be to first compute $QK^\top$ and applying a row-wise top-$k$ operation. Unfortunately, computing $QK^\top \in \mathbb{R}^{L_Q\times L_K}$ explicitly would require $\Omega\left(L_Q\cdot L_K\right)$ memory. We now describe our approach, which avoids this high cost.

\paragraph{Query chunking} 
A simple way to implement attention and reduce its peak memory consumption is to chunk queries: instead of processing all the queries at once, we partition the queries into chunks and process them sequentially, one chunk at a time. For a chunk size $C$, the rows of $Q$ are grouped into $L_Q / C$ contiguous chunks of size $C$ and the attention function (Eq. \ref{eqn:attention}, \ref{eqn:ff}, \ref{eqn:topk}) is computed using $Q_\mathcal{C}, K, V$ as inputs where $Q_\mathcal{C}$ denotes the subset of $Q$ corresponding to chunk $\mathcal{C}$. 

During \emph{inference}, once a query chunk is fully processed, the intermediate activations produced during its processing can be discarded and, hence, the peak memory required to process all $L_Q$ queries is bounded by the memory required to process a single chunk. Therefore, modulo the storage required for $Q, K, V$ and the outputs themselves, the peak memory usage reduces from $\Omega\left(L_Q\cdot L_K\right)$ to $O(C\cdot L_K)$ which is linear with respect to $L_K$ for a fixed chunk size $C$. 

\begin{SCfigure}[][t]
    \centering
    \includegraphics[scale=0.45]{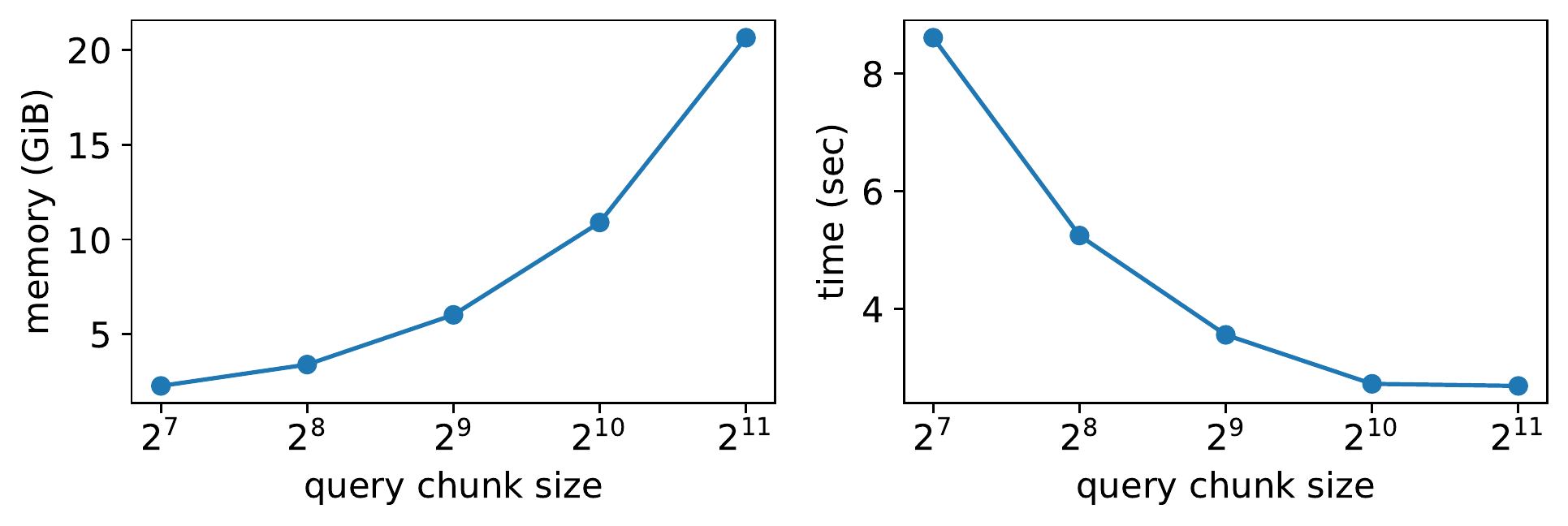}
    \caption{Memory and time required for a forward pass on a single \bert{}-base multi-head self-attention layer on inputs of length $65,536$.}
\label{figure:trade_off}
\end{SCfigure}

Chunk size provides a simple way to trade-off between the maximum memory usage and the slowdown due to the sequential processing of chunks. Fig.~\ref{figure:trade_off} shows memory and time for different chunk sizes for a single BERT-base self-attention layer over a sequence of length $65,536$. We observe that chunk sizes $2^9,2^{10}$ yield a good trade-off between time and memory.

\sidecaptionvpos{figure}{c}

\paragraph{Input checkpointing} While query chunking provides a straightforward approach for bounding the peak memory usage of attention during \emph{inference}, it is not so straightforward to employ it during \textit{training}. Let $d\left(A\right)$ denote the gradient of the loss with respect to a tensor $A$. For a given query chunk $Q_\mathcal{C}$, the intermediate activations produced during the computation of the output $o_\mathcal{C} = \mathrm{Attention}(Q_\mathcal{C},K,V)$ are required for computing $d\left(Q_\mathcal{C}\right)$ from $d\left(o_\mathcal{C}\right)$ via backpropagation. Unfortunately, for the above bound on the peak memory usage to hold, we cannot afford to cache these activations for all the chunks, as done by standard automatic differentiation packages. 

Taking inspiration from \textit{gradient checkpointing} \cite{Chen2016TrainingDN}, we observe that if the inputs $Q_\mathcal{C}, K, V$ are available during the backward pass, we can re-compute $o_\mathcal{C}$ and then use the produced intermediate activations to compute $d\left(Q_\mathcal{C}\right)$ from $d\left(o_\mathcal{C}\right)$. Once $d\left(Q_\mathcal{C}\right)$ is computed, we can again discard the intermediate activations and gradients produced during this step and move on to the next chunk. This ensures that the peak memory usage during the backward pass through the attention layer is bounded by the memory required to backpropagate through a single chunk. 

To summarize, a customized backward pass allows us to utilize query chunking, both during forward and backward passes, and only requires us to cache the inputs to the attention function. For a stack of $N$ attention layers and fixed $d$, this reduces the peak memory usage from $\Omega\left(L_Q\cdot L_K \cdot N\right)$ to $O\left((L_Q + L_K)\cdot N + C\cdot L_K\right)$.  

%We now describe an implementation of top-$k$ attention that combines query chunking, input caching and the top-$k$ operation to obtain memory-efficient top-$k$ attention.

As described above, the combination of query chunking and input checkpointing provides a simple method for reducing the memory-footprint of vanilla attention, independent of top-$k$ attention. Indeed, our benchmarking experiments in \S\ref{sec:benchmark}  demonstrate this. 
However, a drawback of this approach is that, during the backward pass, an implicit second forward pass is performed to re-compute the intermediate activations as described above. This can potentially increase the compute (FLOPs) required for a combined forward and backward pass by $50\%$. We now describe how to further improve both compute and memory by combining query chunking and input checkpointing with top-$k$ attention.

\paragraph{Improving efficiency through top-$k$ attention} We now show that one can avoid re-computing activations in the case of $\mathrm{top}{\text -}k{\text -}\mathrm{Attention}$ (Eq. \ref{eqn:topk}). At a high level, $\mathrm{top}{\text -}k(QK^\top)$ provides a highly compressed but accurate representation of $QK^\top$ and requires only $O(L_Q\cdot k)$ storage, compared to $\Omega(L_Q\cdot L_K)$ for $QK^\top$, where we assume $k \ll L_K$. Hence, we can cache it in addition to $Q, K, V$ without incurring a significant increase in memory usage. 

In the pseudo-code below, we show the forward and backward pass for top-$k$ attention over a query chunk with input checkpointing. The steps of the forward pass that we re-compute during the backward pass is the application of $\mathrm{activation}$ on the output of $\mathrm{top}{\text -}k(QK^\top)$ and forming its matrix representation for subsequent operations (compare Lines 6, 7 and Lines 18, 20). Therefore, the number of FLOPs spent on re-computation in our implementation of top-$k$ attention is at most $O(L_Q\cdot (k+L_K) \cdot N)$, typically much lower than $\Omega(L_Q\cdot L_K \cdot d \cdot N)$, as there is no need to re-compute the dot-products.

%As seen in the below pseudo-code of our implementation, the steps of the forward pass that we re-compute during the backward pass is the application of $\mathrm{activation}$ on the output of $\mathrm{top}{\text -}k(QK^\top)$ and forming its matrix representation for subsequent operations (lines 5, 6). Therefore, the number of FLOPs spent on re-computation during our customized implementation of $\mathrm{top}{\text -}k{\text -}\mathrm{Attention}$ is at most $O(L_Q\cdot (k+L_K) \cdot N)$ which can be significantly less than $\Omega(L_Q\cdot L_K \cdot d \cdot N)$ for the previous case.

Moreover, our benchmarking experiments (\S\ref{sec:benchmark}) show that top-$k$ attention leads to improved memory usage compared to query chunking. This is because in vanilla attention (Eq. \ref{eqn:attention}), while performing the re-computation in the backward pass of a query chunk $Q_\mathcal{C}$, we first re-compute $Q_\mathcal{C}K^\top$, then apply $\mathrm{activation}$ and backpropagate through this operation to compute $d(Q_\mathcal{C}K^\top)$. This implies that at this point there are three $C \times L_K$ matrices in memory. In top-$k$ attention, $\mathrm{activation}$ is applied only on a $C \times k$ matrix and at any given time during the backward pass shown below there is at most one $C \times L_K$ matrix in memory: either $\texttt{actv}$ (Line 20) or $\texttt{d\_dots}$ (Line 23). As our experiments show, this can lead to a much smaller memory footprint for small values of $k, N$.

\comment{
\begin{minted}
[
frame=lines,
framesep=1mm,
baselinestretch=1.1,
fontsize=\footnotesize,
linenos
]
{python}
def forward (Q, K, V, k, activation):
    # Q: query chunk, K: keys, V: values                # [C, d], [L_K, d], [L_K, d]
    dots = matrix_prod(Q, transpose(K))                 # [C, L_K]
    top_dots, top_indices = row_wise_topk(dots, k)      # [C, k], [C, k]
    del dots
    top_actv = activation(top_dots)                     # [C, k]
    actv = matrix(top_actv, top_indices)                
    out = matrix_prod(actv, V)                          # [C, d]
    to_cache(Q, K, V, top_dots, top_indices, activation)
    return out
    
def backward(d_out):
    # d_out: grad of loss w.r.t. out                    # [C, d]
    Q, K, V, top_dots, top_indices, activation = from_cache()
    d_top_actv = matrix_prod(d_out, transpose(V), out_indices=top_indices) # [C, k] 
    # did not cache top_actv so re-compute it to backpropagate
    with compute_grads():
        top_actv = activation(top_dots)                 # [C, k]
    d_top_dots = top_actv.backpropagate(d_top_actv)     # [C, k]
    actv = matrix(top_actv, top_indices)                
    d_V = transpose(matrix_prod(transpose(d_out), actv))# [L_K, d]
    del actv
    d_dots = matrix(d_top_dots, top_indices)            
    d_Q = matrix_prod(d_dots, K)                        # [C, d]
    d_K = transpose(matrix_prod(transpose(Q), d_dots))  # [L_K, d]
    return d_Q, d_K, d_V
\end{minted}
}

\vspace*{10pt}
\begin{lstlisting}[language=iPython]
def forward (Q, K, V, k, activation):
    # Q: query chunk, K: keys, V: values      [C, d], [L_K, d], [L_K, d]
    dots = matrix_prod(Q, transpose(K))                 # [C, L_K]
    top_dots, top_indices = row_wise_topk(dots, k)      # [C, k], [C, k]
    del dots
    top_actv = activation(top_dots)                     # [C, k]
    actv = matrix(top_actv, top_indices)                
    out = matrix_prod(actv, V)                          # [C, d]
    to_cache(Q, K, V, top_dots, top_indices, activation)
    return out
    
def backward(d_out):
    # d_out: grad of loss w.r.t. out                    # [C, d]
    Q, K, V, top_dots, top_indices, activation = from_cache()
    d_top_actv = matrix_prod(d_out, transpose(V), out_indices=top_indices) # [C, k] 
    # did not cache top_actv so re-compute it to backpropagate
    with compute_grads():
        top_actv = activation(top_dots)                 # [C, k]
    d_top_dots = top_actv.backpropagate(d_top_actv)     # [C, k]
    actv = matrix(top_actv, top_indices)                
    d_V = transpose(matrix_prod(transpose(d_out), actv))# [L_K, d]
    del actv
    d_dots = matrix(d_top_dots, top_indices)            
    d_Q = matrix_prod(d_dots, K)                        # [C, d]
    d_K = transpose(matrix_prod(transpose(Q), d_dots))  # [L_K, d]
    return d_Q, d_K, d_V
\end{lstlisting}

\section{Benchmarking}\label{sec:benchmark}

\comment{
\jb{this section dives into technical detail too quickly and doesn't let the reader breath. Should start with something like: `In this section, we compare the memory and runtime of our proposed approach to vanilla transformer and to the Performer, as a representative of state-of-the-art linear attention variants. We benchmark separately (a) a single self-attention layer over long sequences, (b) a single feed-forward layer with a high dimensionality, and (c) a multi-layer transformer with a BERT architecture.' Then you can move on to more technical details. More generally it would be good to maximize interesting content and minimize technical details in the section.}
}

In this section, we benchmark top-$k$ attention in terms of time and memory, and compare it to vanilla attention, query-chunking without the top-$k$ operation, and to Performer \cite{Choromanski2020RethinkingAW}, as a representative of state-of-the-art linear attention variants. We separately benchmark (a) a single self-attention layer over long sequences, (b) a single feed-forward layer with a large feed-forward dimension, and (c) a $12$-layer Transformer decoder with same architecture as \bert{}-base \cite{devlin2018bert}.

\paragraph{Experimental details}
For all models, we benchmark by running a forward and backward pass over random inputs. Each measurement is an average over 3 runs on an Nvidia A100 GPU and is discarded if memory usage exceeds $30$GiB. We use causal masking for self-attention layers to highlight the simplicity of our approach that can seamlessly handle arbitrary attention masks, unlike other methods \cite{Wang2020LinformerSW,katharopoulos2020transformers,Choromanski2020RethinkingAW}, where implementing causal masking requires customized CUDA implementations.
For Performer, we use 256 random features, and the  CUDA implementation from \cite{katharopoulos2020transformers}.

\textbf{Multi-head attention layer}: 
We benchmark a single multi-head attention layer over long sequences in a configuration similar to \bert{}-base:
$d_\text{model}$ is $768$, $12$ heads of size $64$, and feed-forward dimension $3072$. Fig.~\ref{figure:mha} shows the results when setting $k$ to $128$ and the query chunk size to $1024$, which was shown to provide a good time-memory trade-off in \S\ref{subsec:top-k-att}.

We observe that top-$k$ attention has the same device-memory usage as the Performer (top) for sequences as long as $65K$ tokens, while being as fast as vanilla attention, and even faster than Performer on inputs of length up to $4K$. With vanilla attention, we cannot fit even a single multi-head attention layer over a sequence of more than $10K$ tokens, while top-$k$ uses less than $10$GiB of memory over sequences of length $65K$. Lastly, we observe improvement in both time and memory when comparing top-$k$ attention to query chunking over vanilla attention, where using top-$k$ leads to a $3\times$ memory reduction for sequences of length $65K$.

\begin{figure}[t]
    \centering
    \hspace*{-0.17in}
    \subfloat[$12$-layer ``base'' Transformer decoder]{\label{figure:base}\includegraphics[scale=0.48]{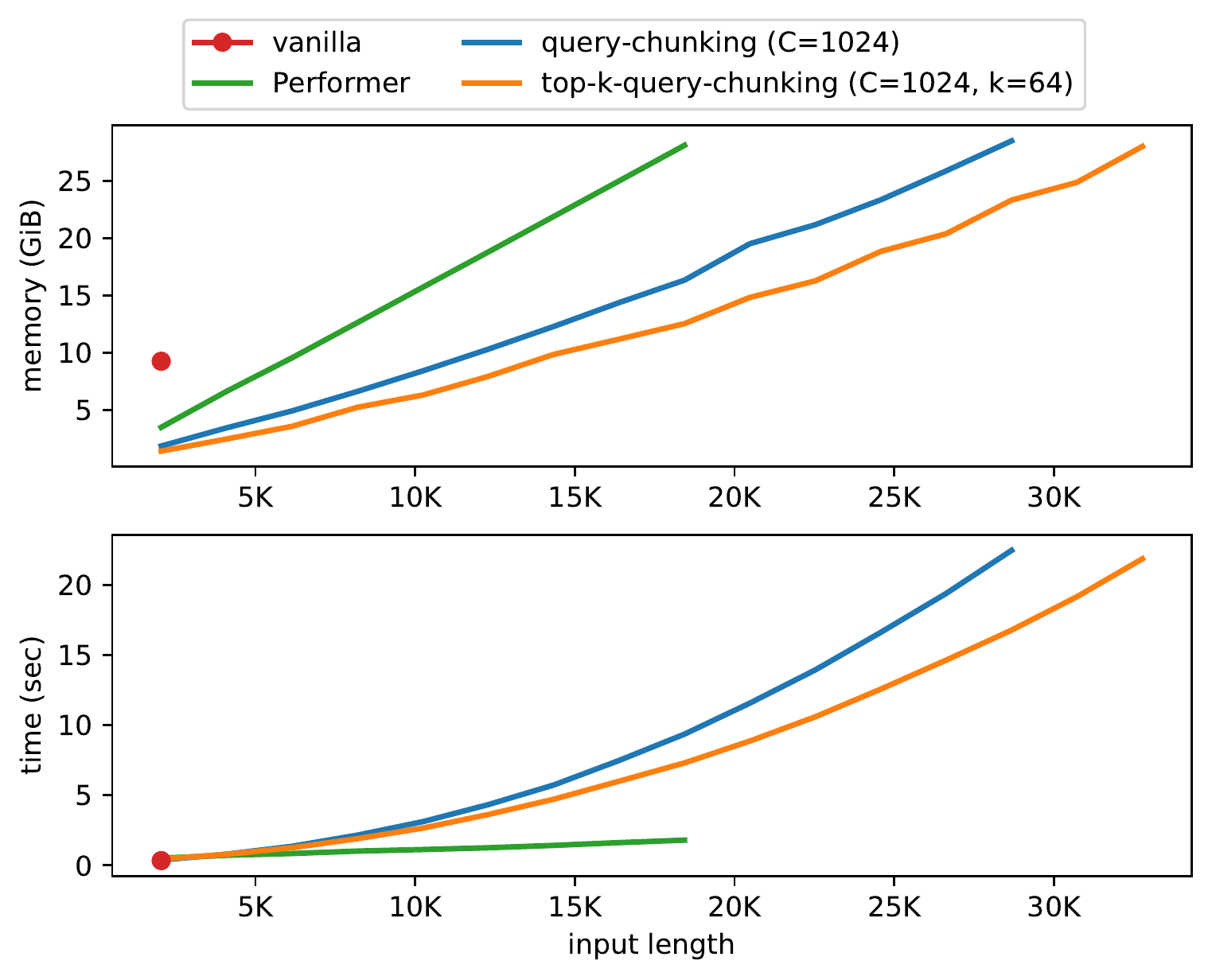}}
    \hspace*{-0.07in}
    \subfloat[single feed-forward layer]{\label{figure:ff}\includegraphics[scale=0.48]{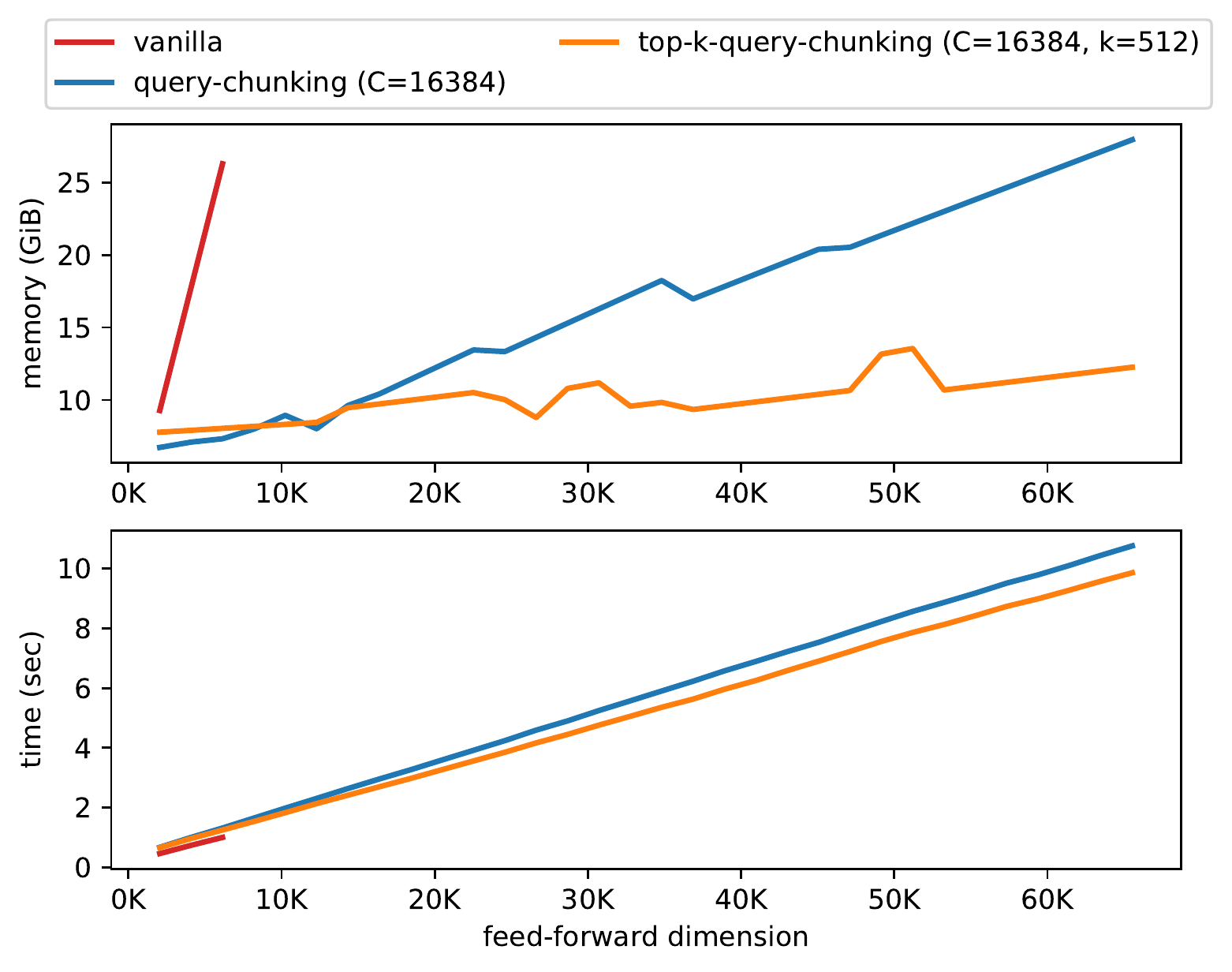}}
    \caption{Memory and time required for a combined forward and backward pass (details in \S\ref{sec:benchmark}).}
\label{figure:benchmark}
\end{figure}

\textbf{Feed-forward layer} : 
While considerable effort has been dedicated to devising efficient models for long contexts, a large feed-forward dimension is useful for knowledge-intensive tasks such as open-domain QA \cite{Roberts2020t5kb,Brown2020LanguageMA}, and efforts have been made to reduce its complexity \cite{fedus2021switch}. 
We benchmark the resource usage of top-$k$ attention at a single feed-forward layer for different feed-forward dimensions using batch size $512$ and input length $512$, which results in $2^{18}$ queries per batch. 

Top-$k$ attention (Figure~\ref{figure:ff}), for $k=512$ and query chunk size $2^{14}$, dramatically improves device-memory usage compared to vanilla attention: it allowed us to use a feed-forward dimension $65K$ with $11$GiB, while vanilla attention uses the same amount of memory with a feed-forward dimension $2K$.
Fitting a linear curve to the memory usage of vanilla attention and top-$k$ attention, we estimate that top-$k$ attention can handle feed-forward dimension $205K$ compared to $7K$ for vanilla attention on a $30$GiB machine. Moreover, comparing top-$k$ attention to query chunking, we again observe a $3\times$ improvement in memory usage when the number of keys is $65K$. Lastly, we observe only a minor slowdown in top-$k$ attention compared to vanilla attention.

%to handle $28\times$ larger feed-forward dimension compared to vanilla attention and $3\times$ larger compared to only chunking, without incurring any significant slowdown \jb{explain the 28x and 3x by saying also the dimension to help look at the figure and understand}. As top-$k$ attention results in a highly sparse matrix of query-key dot products, we also explored the option of using PyTorch's \texttt{torch.sparse} framework for performing sparse $\times$ dense matrix products faster but could not obtain desirable results \jb{I think this can be deleted}. \jb{You have key chunking in the figure but you don't talk about it, so either we drop it or say why it is bad and why we think it is interesting}

\textbf{12-layer model}: 
We benchmark a $12$-layer model to examine the cumulative utility of not caching $QK^\top$ in all $N$ layers compared to the Performer. We use the same architecture as \bert{}-base with batch size $1$ and vary the input length. We use a Transformer decoder with top-$64$ attention and chunk size $1,024$ at the self-attention layers, and simple query chunking with chunk size $4,096$ at the feed-forward layers.

We easily fit a $32K$-long input on a $30$GiB GPU, improving memory consumption by more than $8\times$ compared to vanilla Transformer and $2\times$ compared to Performer. Moreover, top-$k$ attention outperforms query chunking in terms of both memory and runtime. As top-$k$ attention targets memory consumption but not runtime, a current limitation is that runtime, unlike Performer, is still quadratic. Thus, running multi-layer models on long sequences is reasonable in a fine-tuning or zero-shot inference setup, but further work is required for training from scratch 12-layer models over large datasets that contain long sequences.

%For a Transformer decoder (Figure~\ref{figure:base}), using top-$64$ attention at self-attention layers with chunk size $2^{10}$ and query chunking at FF layers with chunk size $2^{12}$ \jb{not top-k at the feed-forward layer, right? this needs to be said I think and say this is BERT-base again} allowed us to easily fit $32K$-long inputs on a $30$GiB GPU, more than $8\times$ that of vanilla attention and $2\times$ the Performer \jb{when you say 8x is it w.r.t input length? be less ambiguous}. 

Overall, our benchmarking results over multi-head attention, feed-forward, and multi-layer Transformer establish top-$k$ attention as a strong baseline for future work on efficient Transformers that dramatically improves memory consumption.
Next, we evaluate top-$k$ attention on downstream tasks and show that top-$k$ attention can be used as a drop-in replacement for vanilla attention without additional pre-training, which can allow resource-constrained research groups experiment with Transformers over long sequences or models with a large feed-forward dimension.

% \ag{say what Q-chunking means in the caption, subset of layers?}
\section{Experimental Evaluation of Top-$k$ Attention}\label{sec:experiments}

Having established top-$k$ attention as a memory efficient alternative to vanilla attention, we now show that, even for small values of $k$, top-$k$ attention provides a high-quality approximation of vanilla attention, both at the multi-head attention and feed-forward layers. We empirically show this in a wide range of setups including (a) training from scratch on tasks that require handling long-range dependencies (\S\ref{sec:lra}) and on language modeling (\S\ref{sec:wiki}), (b) fine-tuning pre-trained language models (\tfive{}) on multiple QA datasets (\S\ref{sec:t5}), and (c) performing zero-shot inference using pre-trained language models (\unifiedqa{}) without any training (\S\ref{sec:unifiedqa}).

\subsection{Long Range Arena}\label{sec:lra}

% \jb{have an opening sentence that helps the reader. Something like: `Long range arena is a recent benchmark that was established to examine the ability of efficient variants of transformers to handle long sequences'}

Long Range Arena \cite{tay2021long} is a recently established benchmark for evaluating the ability of Transformer variants to handle long sequences. It comprises of multiple text classification tasks with inputs containing thousands of tokens (Table \ref{table:lra}). In ListOps \cite{Nangia2018ListOpsAD}, given a sequence of operations on single-digit integers, the model predicts a single-digit solution modeled as $10$-way classification. IMDb movie reviews \cite{maas2011imdb} is a character-level binary sentiment classification task. Lastly, in the ACL Anthology Network (AAN) \cite{Radev2013TheAA} task, a character-level model classifies if there is a citation between a pair of papers.

% \jb{say something about the nature of the task and that the length of sequences in these benchmarks is in  the table?}. 

For each task, we downloaded and directly used the vanilla Transformer code offered by the authors \cite{tay2021long} and compared the performance before and after replacing the multi-head attention layers with top-$128$ attention, using identical hyperparameters for both cases (details in \S\ref{sec:app-lra}).\footnote{\url{https://github.com/google-research/long-range-arena}} 

\begin{figure}[h]
    \centering
    \hspace*{-0.25in}
    \subfloat[ListOps]{\label{figure:listops}\includegraphics[scale=0.37]{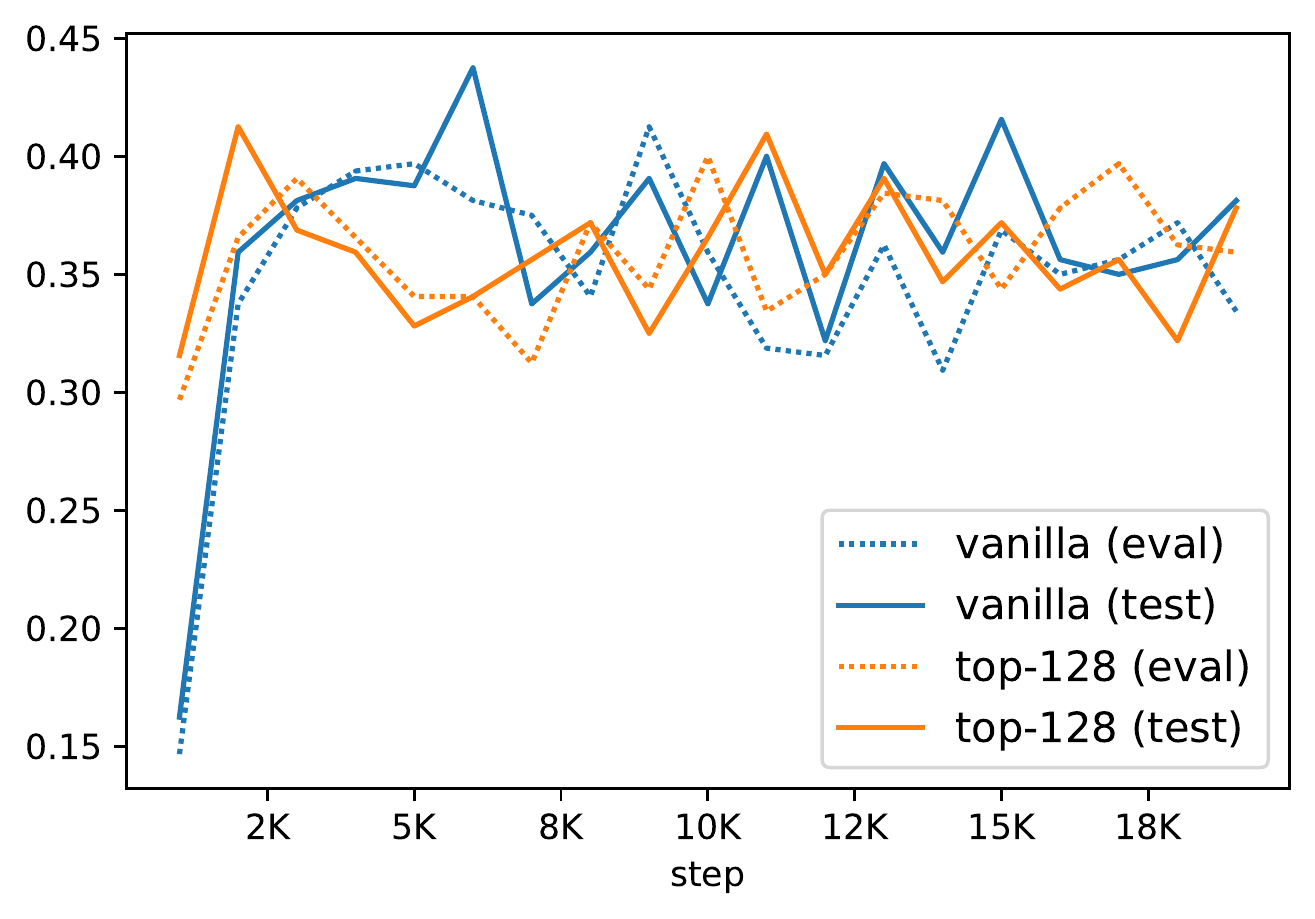}}
    \hspace*{-0.07in}
    \subfloat[IMDb]{\label{figure:imdb}\includegraphics[scale=0.37]{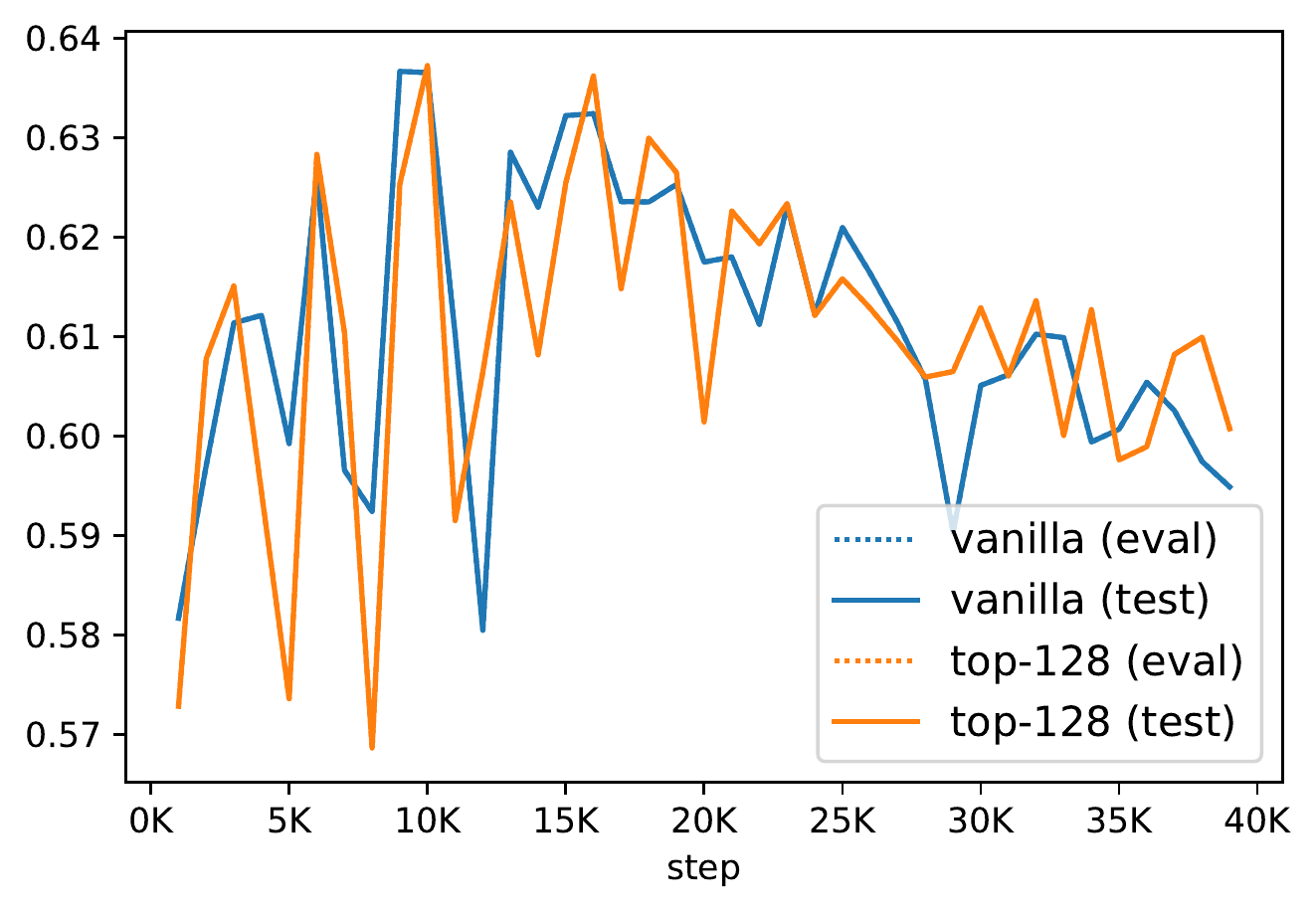}}
    \hspace*{-0.08in}
    \subfloat[AAN]{\label{figure:aan}\includegraphics[scale=0.37]{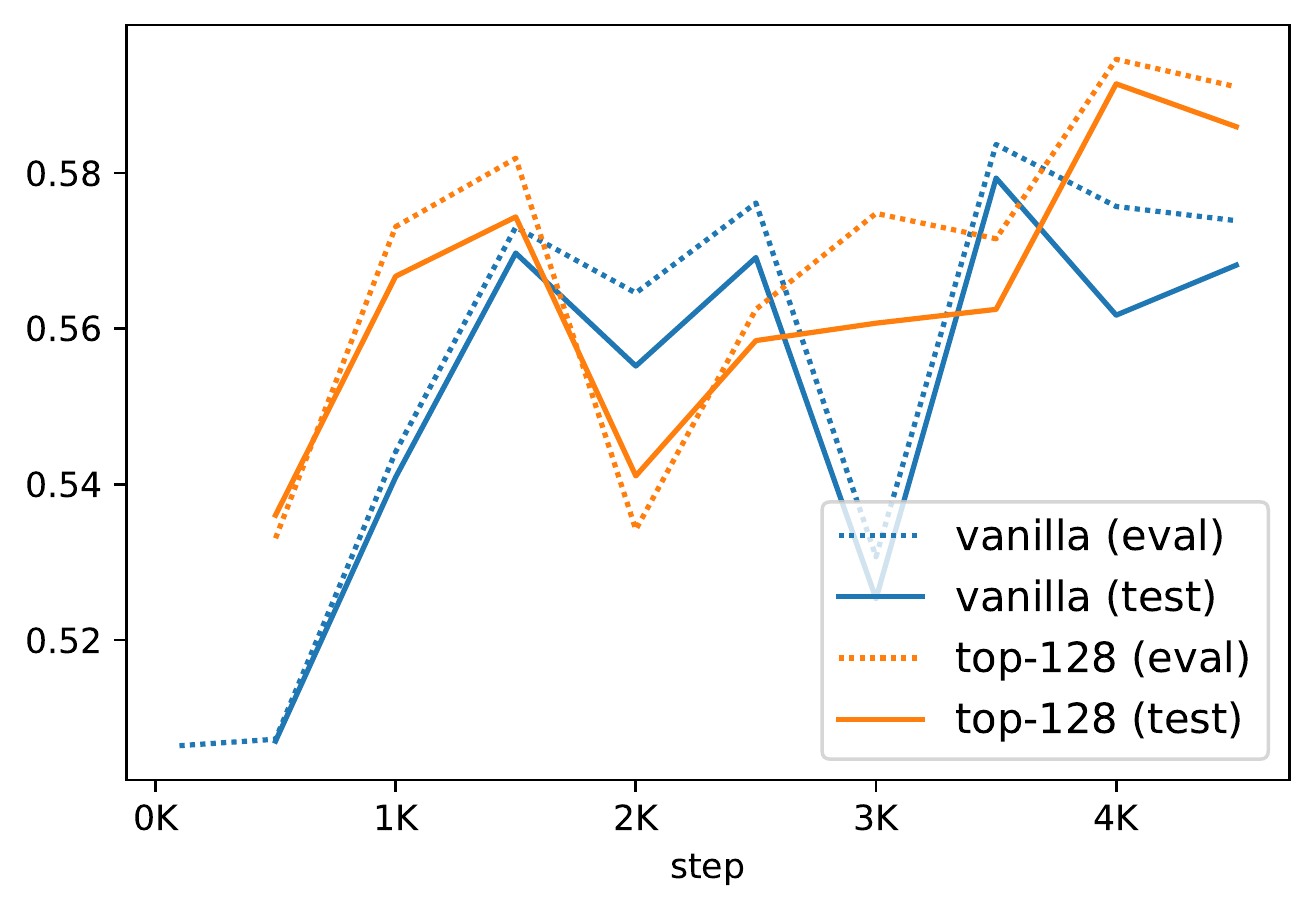}}
    \caption{Learning curves of vanilla and top-$128$ attention on Long Range Arena (\S\ref{sec:lra}).}
\label{figure:lra}
\end{figure}

Test accuracy measured at the training checkpoint with the highest accuracy on the development set is reported in Table \ref{table:lra} and the learning curves on the development and test sets are shown in Fig.~\ref{figure:lra}. On IMDb and AAN, the performance of top-$128$ is comparable or better than vanilla attention. For ListOps, there is a minor drop in performance (1.5 points), but learning curves  (Figure~\ref{figure:listops}) exhibit similar behaviour.

Thus, top-$k$ attention, even for $k$ as small as $3$\% of the number of keys, results in a performance very similar to that of vanilla attention. This shows that an exact and sparse top-$k$ solution is a high-quality approximation for vanilla attention at multi-head attention layers.

\sidecaptionvpos{table}{t}
\begin{SCtable}[][h]
% [h!]\setlength{\tabcolsep}{8pt} 
    \scriptsize
    \centering
    \begin{tabular}{ccccc}\hline
              & ListOps  & IMDb     & AAN   &  mean   \\\hline
    input length & $2K$  & $1K$     & $4K$     \\\hline
    vanilla (reported in \cite{tay2021long}) & $36.4$ & $64.3$ & $57.5$ & $52.7$ \\\hline
    % vanilla (our run)   &  $0.09$  & $63.02$  & $58.00$  \\\hline
    % top-$128$ & $17.65$  & $63.38$  & $57.76$  \\\hline
    vanilla (our run)   &  $\mathbf{38.12}$  & $63.66$  & $57.93$ & $53.2$ \\\hline
    top-$128$           & $36.56$  & $\mathbf{63.72}$  & $\mathbf{59.14}$ &  $53.1$ \\\hline
    \end{tabular}
    \caption{Test accuracy for vanilla and top-$128$ attention on Long Range Arena.}
    \label{table:lra}
\end{SCtable}

\subsection{Language Modeling}\label{sec:wiki}
We further ascertain the findings of \S\ref{sec:lra} via language modeling on WikiText-103 \cite{Merity2017PointerSM} using a $6$-layer Transformer decoder with $156$M parameters. Using an input length of $1024$, we trained two models with vanilla and top-$64$ attentions at the self-attention layers, obtaining test perplexity scores of $30.96$ and $30.51$ respectively, slightly better in case of top-$64$ (details in \S\ref{sec:app-wikitext}).

\subsection{Zero-shot Inference with \unifiedqa{} }\label{sec:unifiedqa}

We have established that the performance of top-$k$ attention is comparable to vanilla attention when training the model from scratch. In this set-up, several recently-proposed approaches have also reported competitive performances \cite{tay2021long}. Now, we consider a different and more practical setup, where the starting point is using an already pre-trained language model \cite{devlin2018bert,raffel2019exploring}.
As such models were trained using vanilla attention, replacing it with a new attention variant typically requires a corrective pre-training stage to allow the model weights to adjust to the new variant, which can be expensive for large models. For example, \cite{gupta2021valueaware,peng2021random} have shown that using random features without corrective pre-training leads to high error rates in a language modeling task. Moreover, as explained in \S\ref{subsec:att_in_transformers}, most past methods are incompatible with feed-forward layers. In the subsequent experiments we show that it is possible to replace vanilla with top-$k$ attention, at multi-head attention and feed-forward layers, and perform inference and fine-tuning without any need for such correction.

First, we compare the performance of \unifiedqa{} \cite{2020unifiedqa} before and after replacing its feed-forward layers with our implementation of top-$k$ attention and directly performing inference on 12 different question answering (QA) datasets without any training. \unifiedqa{} is a \tfive{}-based \cite{raffel2019exploring} model with $11$B parameters \cite{raffel2019exploring}, fine-tuned on a weighted mixture of QA datasets. The 12 datasets include diverse domains, such as science questions, factoid questions over Wikipedia, commonsense questions, etc. Details regarding the datasets and metrics can be found in \S\ref{sec:app-unifiedqa}. %\ag{say how EM was computed} % \jb{add ref to T5}

Table~\ref{table:unifiedqa} shows the results for increasing values of $k$, where the feed-forward dimension of the model is $65,536$. We observe that already when $k=256$ and $k=512$, i.e., less than $1$\% of the number of keys, performance is comparable to vanilla Transformer. When $k=4,096$ ($6$\% of the number of keys), performance is equal or better than vanilla Transformer on all tasks. This highlights the plug-and-play property of top-$k$ attention, which can be used without \emph{any} additional training.
%\footnote{To empirically verify that the Performer indeed requires corrective pre-training, we ran it with UnifiedQA-base on CSQA and obtained an exact match score of $0.0$ compared to $45.3$ with top-$64$ attention.}

%\ag{check} On repeating this experiment with \unifiedqa{}-base (having feed-forward dimension $3072$) using Performer attention (over a large range of number of random features) for the queries, keys and values at the feed-forward layers we got $0$ exact match score on CommonsenseQA compared to $45.1$ for vanilla attention and $45.3$ for top-$64$ attention.

\begin{table}[h!]\setlength{\tabcolsep}{2.3pt} %% default is 6pt
    \scriptsize
    \centering
\begin{tabular}{c|cccccccccccc}
 $k$ & AI2 elem. & AI2 mid. & ARC easy & ARC chal. & BoolQ & CSQA & MCTest & NarQA & OBQA & RACE & SQuAD v1 & SQuAD 2.0 \\\hline
% k & \textbf{} & \textbf{} & \textbf{} & \textbf{} & \textbf{} & \textbf{} & \textbf{} & \textbf{} & \textbf{} & \textbf{} & \textbf{} & \textbf{} \\
64 & 82.9 & 77.6 & 79.3 & 71.6 & 88.7 & 72.4 & 92.8 & 30.5 & 77.4 & 83.5 & 86.1 & 82.0 \\\hline
128 & 87.8 & 81.6 & 83.5 & 72.2 & 89.7 & 74.8 & \textbf{93.1} & 31.6 & 80.0 & 85.8 & 86.8 & 84.3 \\\hline
256 & 91.1 & \textbf{84.0} & 85.3 & 75.6 & 90.0 & 75.9 & 92.8 & 32.1 & \textbf{85.4} & 86.8 & 86.8 & 85.7 \\\hline
512 & 91.1 & 84.0 & 85.4 & 75.9 & 90.4 & 77.0 & 93.1 & \textbf{32.3} & 83.6 & 87.2 & 86.9 & 86.2 \\\hline
1024 & \textbf{91.9} & 83.2 & \textbf{86.3} & \textbf{76.3} & 90.7 & 77.2 & 93.1 & 32.3 & 85.0 & 87.3 & \textbf{87.0} & \textbf{86.3} \\\hline
2048 & 91.9 & 82.4 & 85.8 & 75.9 & \textbf{90.8} & 77.2 & 93.1 & 32.3 & 85.0 & 87.3 & 87.0 & 86.3 \\\hline
4096 & 91.9 & 82.4 & 86.0 & 75.6 & 90.8 & \textbf{77.6} & 93.1 & 32.3 & 85.2 & \textbf{87.4} & 87.0 & 86.3 \\\hline\hline
65536 (vanilla) & 91.9 & 82.4 & 86.0 & 75.6 & 90.7 & 77.5 & 93.1 & 32.3 & 85.2 & 87.4 & 87.0 & 86.3 \\\hline
\end{tabular}
\caption{Exact match scores of \unifiedqa{} on development sets with top-$k$ attention at feed-forward layers. \textit{Notation}: AI2 science elementary (AI2 elem.), AI2 science middle (AI2 mid.), ARC challenging (ARC chal.), CommonsenseQA (CSQA), NarrativeQA (NarQA), OpenbookQA (OBQA).}
\label{table:unifiedqa}
\end{table}

\subsection{Zero-shot Inference with \bert{}}\label{sec:squad}
 
To verify that the plug-and-play property of top-$k$ attention also holds at self-attention layers, we downloaded a \bert{}-large-uncased-whole-word-masking checkpoint \cite{devlin2018bert} already fine-tuned on SQuAD v1 \cite{rajpurkar2016squad} and evaluated its performance on the development set before and after replacing its self-attention layers with top-$k$ attention. For $k$ as low as $16$ ($4$\% of input length), we only saw a minor decrease in the exact match scores ($86.9 \rightarrow 86.2$). Moreover, to empirically verify that dense approximations of vanilla attention (Performer, RFA, etc) indeed require corrective pre-training, we repeated the measurement using Performer attention with $256$ features, obtaining a score of $0.38$.

\subsection{\tfive{} Finetuning}\label{sec:t5}

Having established the plug-and-play property of top-$k$ attention in zero-shot inference (\S\ref{sec:unifiedqa}, \S\ref{sec:squad}), we now show the effectiveness of top-$k$ attention when fine-tuning a model, and that there are no unforeseen issues stemming from training under high sprasity. Here, we use T5-base rather than T5-11B and evaluate on five QA datasets (and not 12) due to computational constraints.

Similar to \S\ref{sec:unifiedqa}, we replace the feed-forward layers of \tfive{}-base, which has  feed-forward dimension $3072$, with our implementation of top-$256$ attention and fine-tuned on multiple QA datasets. As summarized in Table \ref{table:t5}, we found that the performance of top-$256$ attention was again comparable to vanilla attention on BoolQ, CommonsenseQA and ROPES with a minor loss in performance on MCTest ($81.2 \rightarrow 79.4$) and OpenbookQA ($58.8 \rightarrow 58.0$).

\sidecaptionvpos{table}{t}
\begin{SCtable}[70][h]\setlength{\tabcolsep}{2pt}
% \begin{table}[h!]\setlength{\tabcolsep}{2pt} %% default is 6pt
    \scriptsize
    \centering
\begin{tabular}{cccccc}
                       & BoolQ & CSQA & MCTest & OBQA & ROPES \\\hline
T5-base & 0.5 & 35.7 & 36.6 & 17.0 & 21.7 \\\hline
\unifiedqa{}-base & 82.0 & 45.0 & 85.3 & 59.6 & 26.3 \\\hline
% \unifiedqa{}-base, top-$256$ & 82.0 & 45.0 & 85.3 & 59.6 & 26.2 \\\hline
T5-base + finetuning & 83.3 & 61.9 & 81.2 & 58.8 & 54.0 \\\hline
T5-base, top-$256$ + finetuning & 83.1 & 62.0 & 79.4 & 58.0 & 53.8 \\\hline
\end{tabular}
\caption{Exact match scores on development sets. ``Finetuning'' denotes model was finetuned on the dataset, else it was evaluated directly without any training. All models use vanilla feed-forward layers except the ones that say top-$256$ (\S\ref{sec:t5}).}
\label{table:t5}
\end{SCtable}

% \begin{table}[h!]\setlength{\tabcolsep}{2pt} %% default is 6pt
%     \scriptsize
%     \centering
% \begin{tabular}{cccccc}
%                       & BoolQ & CSQA & MCTest & OBQA & ROPES \\\hline
% T5-base & 0.5 & 35.7 & 36.6 & 17.0 & 21.7 \\\hline
% \unifiedqa{}-base & 82.0 & 45.0 & 85.3 & 59.6 & 26.3 \\\hline
% % \unifiedqa{}-base, top-$256$ & 82.0 & 45.0 & 85.3 & 59.6 & 26.2 \\\hline
% T5-base + finetuning & 83.3 & 61.9 & 81.2 & 58.8 & 54.0 \\\hline
% T5-base, top-$256$ + finetuning & 83.1 & 62.0 & 79.4 & 58.0 & 53.8 \\\hline
% \end{tabular}
% \caption{Exact match scores on development sets. ``Finetuning'' denotes the model was finetuned on the dataset, otherwise it was evaluated directly without any training. All models use vanilla feed-forward layers except the ones that say top-$256$ (\S\ref{sec:t5}). \textit{Notation}: CommonsenseQA (CSQA), OpenbookQA (OBQA).}
% \label{table:t5}
% \end{table}

To summarize, our experiments in \S\ref{sec:lra}-\S\ref{sec:t5} demonstrated that the performance of vanilla attention and top-$k$ attention is comparable at both multi-head attention (\S\ref{sec:lra}, \S\ref{sec:squad}) and feed-forward layers in multiple set-ups including training from scratch (\S\ref{sec:lra}, \S\ref{sec:wiki}), fine-tuning (\S\ref{sec:t5}) and zero-shot inference (\S\ref{sec:unifiedqa}, \S\ref{sec:squad}), while dramatically improving memory usage, as shown in \S\ref{sec:benchmark}.

\section{Discussion}\label{sec:limits}

\textbf{Related work}\ \ \ Our work follows a long line of works on efficient Transformers (see \S\ref{sec:intro}). Our method employs three main ideas: (a) computing the top-$k$ attention scores for each query (b) grouping the queries into chunks and processing these sequentially (c) caching only a part of the activations for the backward pass. Top-$k$ operation was used at self-attention layers by \cite{zhao2020sparse} to show improved model performance, attributed to the removal of irrelevant information in the context. We use it to reduce the resource usage of multi-head attention and feed-forward layers. Processing query chunks sequentially was also used in Reformer \cite{kitaev2020reformer} as activations are not cached. But in that case, by replacing vanilla residual connections in the Transformer with reversible connections \cite{GomezRUG17}. Similar to the explanation provided in \S\ref{subsec:top-k-att}, these require an extra implicit forward pass during the backward pass and do not provide the compute and memory savings we get from our top-$k$ specific backward pass (\S\ref{subsec:top-k-att}). Secondly, replacing residual connections with reversible ones changes the function computed by the model and would require corrective pre-training to be used with \bert{}, \tfive{}, etc (\S\ref{sec:unifiedqa}-\S\ref{sec:t5}).

\textbf{Limitations and future work}\ \ \ As our method requires computing inner products of all queries and keys, it has a quadratic compute requirement. As seen in our pseudo-code (\S\ref{subsec:top-k-att}), there are \textit{four} matrix products (Lines 8, 15, 21, 25) involving a large sparse matrix and a small dense one. Our current implementation does not leverage this sparsity and hence is as slow as vanilla attention. While future devices might allow faster sparse-dense products, in the immediate future, one can leverage block-sparse kernels \cite{child2019generating,Tillet2019TritonAI} which have been successfully used for such products \cite{Rasley2020DeepSpeedSO}.

\textbf{Conclusion}\ \ \ In this work, we proposed a memory-efficient and accurate sparse approximation of the primary sub-layers of a Transformer, benchmarked the resulting resource savings, and verified its quality and unique advantages, on a wide range of downstream tasks and evaluation set-ups.

\begin{ack}
We thank Achille Fokoue, Maria Chang and Avi Sil for helpful discussions. Most of our experiments were conducted on IBM's Cognitive Computing Cluster with additional resources from Hybrid Cloud Infrastructure Research. This research was partially supported by (1) The Yandex Initiative for Machine Learning, and the European Research Council (ERC) under the European Union Horizons 2020 research and innovation programme (grant ERC DELPHI 802800), and (2) the IBM AI Residency program.
\end{ack}

\small
\bibliography{all}
\bibliographystyle{abbrv}
\normalsize

\clearpage

\appendix

\section{Supplemental Material}\label{sec:supplemental}

\subsection{Details of Long Range Arena}\label{sec:app-lra}

ListOps \cite{Nangia2018ListOpsAD} aims to diagnose the capability of modelling hierarchically structured data. Given a sequence of operations on single-digit integers, the model predicts the solution, also a single-digit integer modeled as a $10$-way classification. Character-level text classification with the IMDb movie review dataset \cite{maas2011imdb} is a binary sentiment classification task. In the character-level document retrieval with the ACL Anthology Network (AAN) \cite{Radev2013TheAA}, the model classifies if there is a citation between a pair of papers. 

We used the code and pre-processed data provided by the authors of Long Range Arena \cite{tay2021long} and default model configurations. For each task, we used identical hyperparameters for vanilla and top-$k$ attentions (Table \ref{table:hyperparams-lra}) and used at most two Nvidia A100 for each run.

\textbf{Notation}\ \  BSZ: effective batch size, SQL: input sequence length, LR: learning rate, WRM: linear LR warm-up steps, STEP: number of gradient updates, EFQ: evaluated every these many steps, NL: number of layers in encoder/decoder, HS: hidden size, FF: feed-forward dimension, NH: number of heads, VOC: vocabulary size, DRP: dropout rate, CLIP: maximum gradient norm.

\begin{table}[h!]
    \scriptsize
    \centering
    \begin{tabular}{cccccccccccc} \hline
        task &   BSZ & SQL & LR & WRM & STEP & EFQ & NL & HS & FF & NH  \\\hline
        ListOps & 16 & 2K & 0.02 & 1K & 20K & 1K & 6 & 512 & 2048 & 8 \\\hline
        IMDb &   32 & 1K & 0.03 & 8K & 40K & 1K & 4 & 256 & 1024 & 4  \\\hline
        AAN &   8 & 4K & 0.05 & 8K & 5K &  500 & 4 & 128 & 512 & 4  \\\hline
    \end{tabular}
    \caption{Hyperparameters for LRA tasks (\S\ref{sec:lra}). Other hyperparameters were used as provided at \url{https://github.com/google-research/long-range-arena}.}
    \label{table:hyperparams-lra}
\end{table}

\subsection{Details of \unifiedqa{} inference \& T5 finetuning}\label{sec:app-unifiedqa}
We used Hugging Face's Transformers library \cite{Wolf2019HuggingFacesTS} for these experiments. Authors of \unifiedqa{} collected and pre-processed several QA datasets into a common format: ``{\tt QUESTION \textbackslash n CHOICES \textbackslash n CONTEXT}''. We downloaded this data by following the instructions provided by the authors\footnote{\url{https://github.com/allenai/unifiedqa}} and used it for the \unifiedqa{} inference experiments (\S\ref{sec:unifiedqa}). Some statistics are shown in Table \ref{table:unifiedqa-datasets}. Longer inputs were truncated to 512 tokens.

Given an instance from the pre-processed data, we computed the exact match score of a prediction with respect to the list of provided answers via the SQuAD v1 evaluation script \cite{rajpurkar2016squad}.

% \comment{
\begin{SCtable}[][h!]
    \scriptsize
    \centering
    \begin{tabular}{cccccccccccc} \hline
         &  & training samples & eval samples & \begin{tabular}[c]{@{}c@{}} \tfive{} tokens per sample \\ (90th percentile)  \end{tabular}    \\\hline
        CSQA  &   \cite{Talmor2019CommonsenseQAAQ}     &  9741 & 1221 & 54     \\\hline
        OBQA &   \cite{Mihaylov2018CanAS}     & 4957  & 500  & 60     \\\hline
        ARC easy &  \cite{Clark2018ThinkYH}      & 2251  & 570     & 90     \\\hline
        AI2 elem. &  \cite{Clark2015ElementarySS}   &  623   &   123    &  96      \\\hline
        ARC chal. & \cite{Clark2018ThinkYH}       & 1119  & 299     & 101     \\\hline
        BoolQ & \cite{clark-etal-2019-boolq}       & 9427  & 3270  & 253     \\\hline
        SQuAD 2.0 &  \cite{rajpurkar2018squad}      & 130124  & 11873      & 294     \\\hline
        SQuAD v1 &  \cite{rajpurkar2016squad}      & 87489  & 10570     & 295     \\\hline
        ROPES &   \cite{Lin2019ReasoningOP}     & 10924  & 1688     & 345     \\\hline
        MCTest &    \cite{Richardson2013MCTestAC}    & 1480  & 320     & 435     \\\hline
        RACE &    \cite{Lai2017RACELR}    & 87860  & 4887    & 568     \\\hline
        NarQA & \cite{Kocisk2018TheNR}       & 65494  & 6922    & 1278     \\\hline
    \end{tabular}
    \caption{Statistics of the pre-processed datasets used in \S\ref{sec:unifiedqa} and \S\ref{sec:t5}.}
    \label{table:unifiedqa-datasets}
\end{SCtable}
% }

\comment{
\begin{SCtable}[][h!] 
    \scriptsize
    \centering
    \begin{tabular}{cccccccccccc} \hline
         &  training samples & eval samples & \begin{tabular}[c]{@{}c@{}} \tfive{} tokens per sample \\ (90th percentile)  \end{tabular}    \\\hline
        CSQA      &  9741 & 1221 & 54     \\\hline
        OBQA     & 4957 &  500  & 60     \\\hline
        ARC easy     & 2251 & 570     & 90     \\\hline
        AI2 elem.   &  623   &   123    &  96      \\\hline
        ARC chal.      & 1119 & 299     & 101     \\\hline
        BoolQ       & 9427 & 3270  & 253     \\\hline
        SQuAD 2.0     & 130124 & 11873      & 294     \\\hline
        SQuAD v1     & 87489 & 10570     & 295     \\\hline
        ROPES     & 10924 & 1688     & 345     \\\hline
        MCTest    & 1480 & 320     & 435     \\\hline
        RACE    & 87860 &  4887    & 568     \\\hline
        NarQA       & 65494 &  6922    & 1278     \\\hline
    \end{tabular}
    \caption{Statistics of the pre-processed datasets used in \S\ref{sec:unifiedqa} and \S\ref{sec:t5}.}
    \label{table:unifiedqa-datasets}
\end{SCtable}
}

For the \tfive{} experiments (\S\ref{sec:t5}), we used a slightly different input format. Given an instance in the \unifiedqa{} format, we formed the modified instance as ``{\tt question: QUESTION  context: <yes> <no> <No Answer>  CHOICES \textbackslash n CONTEXT}''.
\begin{SCtable}[][h!]
    \scriptsize
    \centering
    \begin{tabular}{cccccccccccc} \hline
        dataset &   BSZ & LR  & STEP & SQL & CLIP  \\\hline
         CSQA   &   576 & 5e-5 & 3K  & 512 & 1     \\\hline
         OBQA   &   512 & 5e-5 & 3K  & 512 & 1     \\\hline
         BoolQ  &   80 & 5e-5 & 3K  & 512 & 1     \\\hline
         MCTest &   80 & 5e-5 & 3K  & 512 & 1     \\\hline
         ROPES &    80 & 5e-5 & 3K  & 512 & 1     \\\hline
    \end{tabular}
    \caption{Hyperparameters for \tfive{}-base finetuning (\S\ref{sec:t5}). Trainings were performed on a single Nvidia V100 using Adam optimizer with a constant LR and evaluation was done only at end of training.}
    \label{table:hyperparams-t5}
\end{SCtable}

\subsection{Language Modeling on WikiText-103}\label{sec:app-wikitext}
WikiText-$103$ is a language modeling task based on English Wikipedia. We used the language modeling framework provided by Faiseq\footnote{\url{https://github.com/pytorch/fairseq}} and hyperparameters in Table \ref{table:hyperparams-wiki}. The details of Adam optimizer are $\beta_1$=$0.9$, $\beta_2$=$0.98$, weight-decay: $0.01$, CLIP: none, LR schedule: inverse square root. During evaluation on test set, dataset is chunked into segments of length $1024$ and perplexity is computed over each segment normally without access to other segments. 

\begin{table}[h!]
    \scriptsize
    \centering
    \begin{tabular}{cccccccccccc} \hline
        BSZ & SQL & LR & WRM & STEP & & NL & HS & FF & NH  & VOC  & DRP \\\hline
        64 & 1024 & 5e-4 & 4K & 50K & & 6 & 512 & 2048 & 8 & 267744 & 0.1 \\\hline
    \end{tabular}
    \caption{Hyperparameters for WikiText-103 task (\S\ref{sec:wiki}).}
    \label{table:hyperparams-wiki}
\end{table}

\subsection{Benchmarking details}\label{sec:app-benchmarking}
Benchmarking (\S\ref{sec:benchmark}) was done in PyTorch 1.8.1. For each run, we sampled a batch of random 32-bit input vectors and a backward pass was performed using the mean of the output elements as the loss. The part of code that was timed was enclosed within {\tt torch.cuda.synchronize()} to ensure all CUDA threads finished. Memory usage was measured using {\tt  torch.cuda.max\_memory\_reserved()}. On Nvidia A100, any internal casting to \textrm{TF32} was explicitly disabled.

\begin{SCfigure}[][h!]
    \centering
    \includegraphics[scale=0.5]{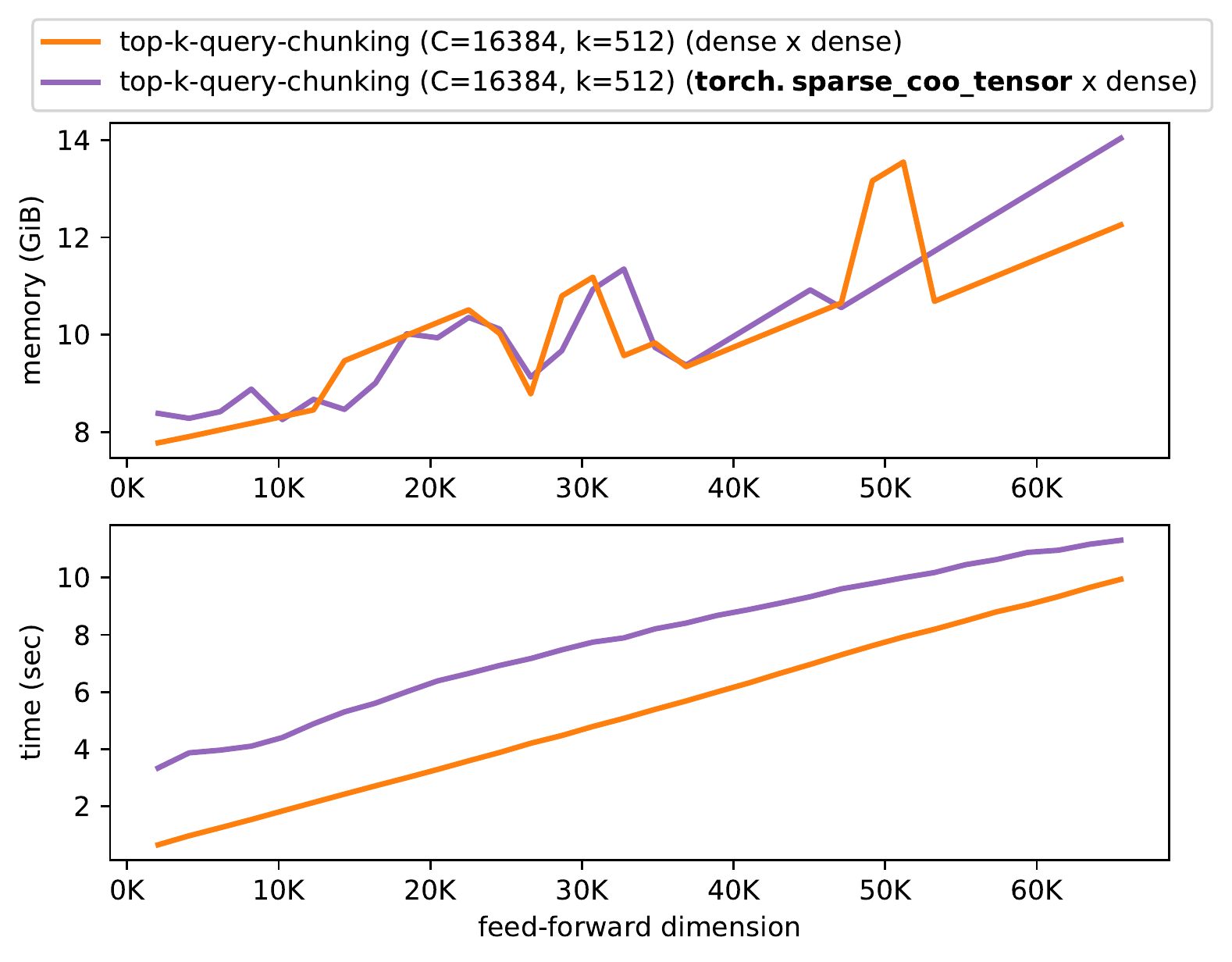}
    \caption{Memory and time required for a combined forward and backward pass on a single feed-forward layer using random inputs.}
\label{figure:supp-ff-sparse}
\end{SCfigure}

We considered the option of performing matrix products involving large sparse matrices (Lines 8, 15, 21, 25 in our pseudo-code (§2.2)) by representing them in {\tt torch.sparse\_coo\_tensor} format and using the {\tt torch.sparse} framework to explicitly leverage the sparsity. Unfortunately, we could not obtain encouraging results even for $k =$ 1\% of number of keys (Figure \ref{figure:supp-ff-sparse}) and plan on experimenting with block-sparse kernels \cite{Tillet2019TritonAI} in the near future.

\sidecaptionvpos{figure}{c}

\end{document}